\documentclass[10pt]{article}
\usepackage{style}

\title{Pretraining with random noise for uncertainty calibration}
\author{Jeonghwan Cheon and Se-Bum Paik$^{\text{*}}$ \\
Department of Brain and Cognitive Sciences, Korea Advanced Institute of Science and Technology, Daejeon 34141, Republic of Korea \\
* Correspondence author: sbpaik@kaist.ac.kr}

\begin{document}
\maketitle

\section*{Abstract}
Uncertainty calibration is crucial for various machine learning applications, yet it remains challenging. Many models exhibit hallucinations — confident yet inaccurate responses — due to miscalibrated confidence. Here, we show that the common practice of random initialization in deep learning, often considered a standard technique, is an underlying cause of this miscalibration, leading to excessively high confidence in untrained networks. Our method, inspired by developmental neuroscience, addresses this issue by simply pretraining networks with random noise and labels, reducing overconfidence and bringing initial confidence levels closer to chance. This ensures optimal calibration, aligning confidence with accuracy during subsequent data training, without the need for additional pre- or post-processing. Pre-calibrated networks excel at identifying “unknown data,” showing low confidence for out-of-distribution inputs, thereby resolving confidence miscalibration.

\renewcommand{\figurename}{Fig.}
\vspace{0.5cm}
\section{Introduction}

\begin{figure}[t!]
    \centering
    \includegraphics[width=120mm]{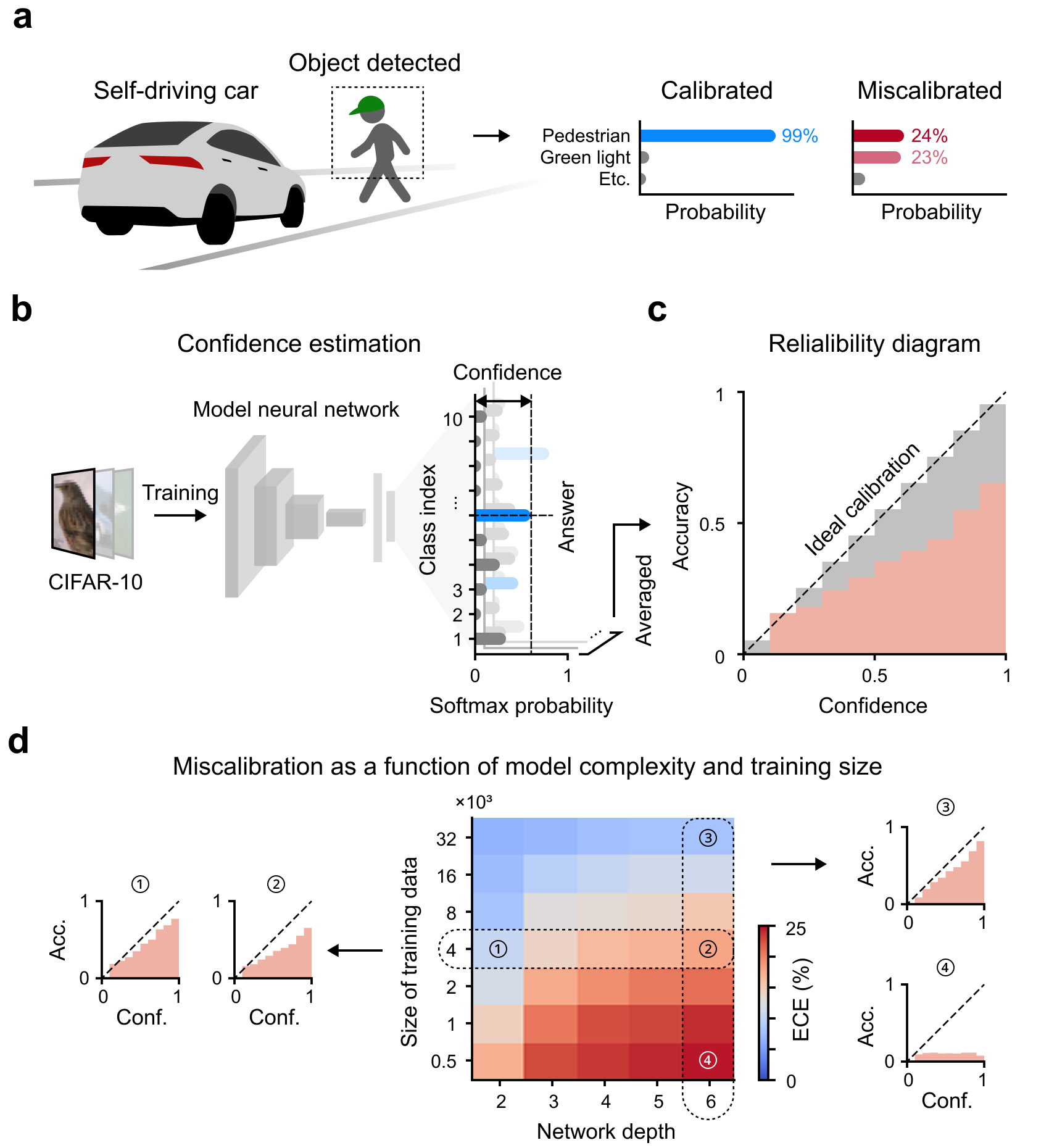}
    \caption{
        \textbf{Confidence miscalibration in artificial neural networks.} \\
        (a) The illustration depicts how self-driving cars detect objects in their environment and make decisions based on these detections. Both the calibrated and miscalibrated models predict the same label, but the confidence levels differ significantly between the two models. Miscalibrated confidence can lead to incorrect decisions, even when the model’s accuracy is high.
        (b) The predicted answer and its confidence are calculated using the probability output from the SoftMax function.
        (c) Reliability diagram for a six-layer feedforward neural network trained on a subset of the CIFAR-10 dataset (training data size = 4,000). The test dataset's predictive confidence and correctness are binned based on confidence values, and the accuracy is calculated in each bin. The diagonal line represents ideal calibration, where confidence perfectly matches the expected accuracy. The Expected Calibration Error (ECE) is the difference between predicted confidence and actual accuracy.
        (d) Calibration error across various model complexities and training data sizes. Each color represents the ECE: blue and red indicate low and high calibration error, respectively.
    }
    \label{fig1}
\end{figure}

Recent advances in deep neural networks have significantly improved their ability to learn with high accuracy \cite{lecun2015}. For example, deep learning models achieved human-level accuracy in classifying natural images over a decade ago \cite{lecun1998, Krizhevsky2012, he2016, huang2017}. These models are now being applied to a wide range of real-world problems that involve complex decision-making, such as autonomous driving \cite{chen2017}, medical diagnosis \cite{shen2017}, and financial engineering \cite{fischer2018}. In these applications, simply inferring the correct answer is insufficient; an additional dimension of information, confidence or uncertainty, is crucial to predict answers with calibrated probabilities that accurately reflect the likelihood of correctness (Fig. \ref{fig1}a). However, modern deep neural networks are often poorly calibrated for confidence estimation \cite{guo2017, nalisnick2019, ovadia2019}. Ironically, the increase in model capacity enables accurate performance but often negatively impacts confidence calibration \cite{guo2017}. Specifically, many recent models frequently exhibit overconfidence, even with out-of-distribution samples, despite lacking knowledge about them \cite{hendrycks2017, liang2018}. This miscalibration leads to poor decision-making in real-world applications \cite{helldin2013, mehrtash2020} — For example, hallucinations in large language models \cite{achiam2023, team2023}, where the model confidently produces false and unsubstantiated outputs \cite{farquhar2024, xiao2021}, are a well-known issue resulting from the miscalibration of neural networks \cite{geng2024, groot2024}.

Previous studies have tried to address calibration issues through additional pre- or post-processing techniques. However, these methods require additional, computationally expensive steps to obtain calibrated confidence \cite{guo2017, niculescu2005, platt1999, zadrozny2001, zadrozny2002}. Many state-of-the-art algorithms employ auxiliary networks or score functions to differentiate between overconfident predictions for unseen out-of-distribution data and those for in-distribution data \cite{hendrycks2018, lee2018b, liu2020}. In general, the fundamental problem is that previous approaches typically treat in-distribution and out-of-distribution data separately, which essentially recalibrates the confidence values through separate calculation processes to achieve the desired results.

We can ask a more fundamental question: What causes uncertainty miscalibration in deep neural networks? Unlike machines, human intelligence naturally exhibits metacognition, enabling the estimation of uncertainty and awareness of what is “known” and “unknown” \cite{cosmides1996, fleming2012, kepecs2008, kiani2009, masset2020}. So, what is the underlying mechanism of uncertainty calibration in the brain, and how can artificial neural networks learn brain-like uncertainty calibration, possibly without post-processing or auxiliary computation? To address this, we draw inspiration from our previous work \cite{cheon2024}, which proposed pretraining with random noise as a method that mimics the biological prenatal learning stage. Specifically, we hypothesize that these initial pre-calibrations could facilitate the learning of calibrated uncertainty.

In this study, we demonstrate that the commonly used method of random initialization in deep learning, long regarded as a standard technique, is, in fact, the underlying cause of network uncertainty miscalibration. The solution to this miscalibration issue is achieved simply through a novel initialization strategy that deviates from conventional practices — by pretraining networks using random noise and labels, without additional pre- or post-processing methods. Our findings show that randomly initialized neural networks often exhibit high confidence even when they lack sufficient knowledge, but training with random noise effectively calibrates this confidence to uniform chance levels across the input spaces. As a result, the networks learn calibrated probabilities, with accuracy and confidence aligned throughout subsequent data training. Additionally, pre-calibrated networks show lower confidence when encountering unknown data, which helps in detecting out-of-distribution samples. Overall, our results suggest that pretraining with random noise is a simple yet powerful strategy for addressing uncertainty calibration issues.

\section{Results}

\subsection{Failure of confidence calibration in deep neural networks}

To investigate the pattern of confidence miscalibration in deep neural network models, we first used a feedforward neural network designed for pattern classification of three-channel natural images (32×32×3) into ten categories (Fig. \ref{fig1}b) (see Methods for details). The network consists of multiple hidden layers with ReLU nonlinearities and a final classification layer that employs the SoftMax function. For a given image sample, the model outputs probabilities for each object category, and the predicted label corresponds to the category with the maximum probability. The model also provides a measure of prediction confidence, which is derived from the probability assigned to the predicted category. The predictive uncertainty can be quantified as the difference between the confidence value and one. Ideally, a neural network is expected to output calibrated confidence, where the confidence level accurately reflects the actual likelihood of the prediction being correct (Fig. \ref{fig1}c and Supplementary Fig. \ref{figs1}). In contrast, a poorly calibrated network may output confidence levels that do not correspond to the actual likelihood of correctness, leading to potential misinterpretations of the model’s predictions.

We trained the feedforward neural network using a subset of the CIFAR-10 dataset \cite{krizhevsky2009}, which consists of natural images labeled with ten classes representing animals and objects. After training, we evaluated the model's predictions and confidence using test images that were not part of the training set. Specifically, we investigated whether the confidence level of the trained network accurately reflects the likelihood of correctness (Supplementary Fig. \ref{figs1}). By measuring binned confidence values and the corresponding expected accuracy within each binned trial prediction, we constructed a reliability diagram \cite{degroot1983, niculescu2005} to visualize the model's calibration (Fig. \ref{fig1}c). In this diagram, ideal calibration appears as an identity function, where the estimated confidence corresponds to the actual probability of correctness. 

We investigated the pattern of miscalibration by training neural networks of varying complexity with different training data sizes (Fig. \ref{fig1}d). To quantitatively measure the degree of miscalibration, we calculated the Expected Calibration Error (ECE) \cite{naeini2015}, which averages the difference between accuracy and confidence in each bin, weighted by the number of samples. In general, we observed a significant gap between ideal calibration and the model's outputs — the model's accuracy was lower than its confidence level, demonstrating the model's tendency for overconfidence in most conditions. We found that both model complexity and training data size systematically affect confidence calibration. Specifically, the degree of miscalibration increases as the training data size decreases and as model complexity increases, suggesting that miscalibration occurs when the training data is insufficient relative to the network's complexity. This result explains why state-of-the-art models with high complexity struggle with the mismatch between confidence and accuracy. Recent network architectures often increase in depth to enhance learning capacity, but the available data size does not scale accordingly. In real-world applications, acquiring large datasets is costly and computationally demanding, making miscalibration inevitable in deep learning models.

\subsection{Pretraining with random noise for confidence calibration}

\begin{figure}[p!]
    \centering
    \includegraphics[width=\textwidth]{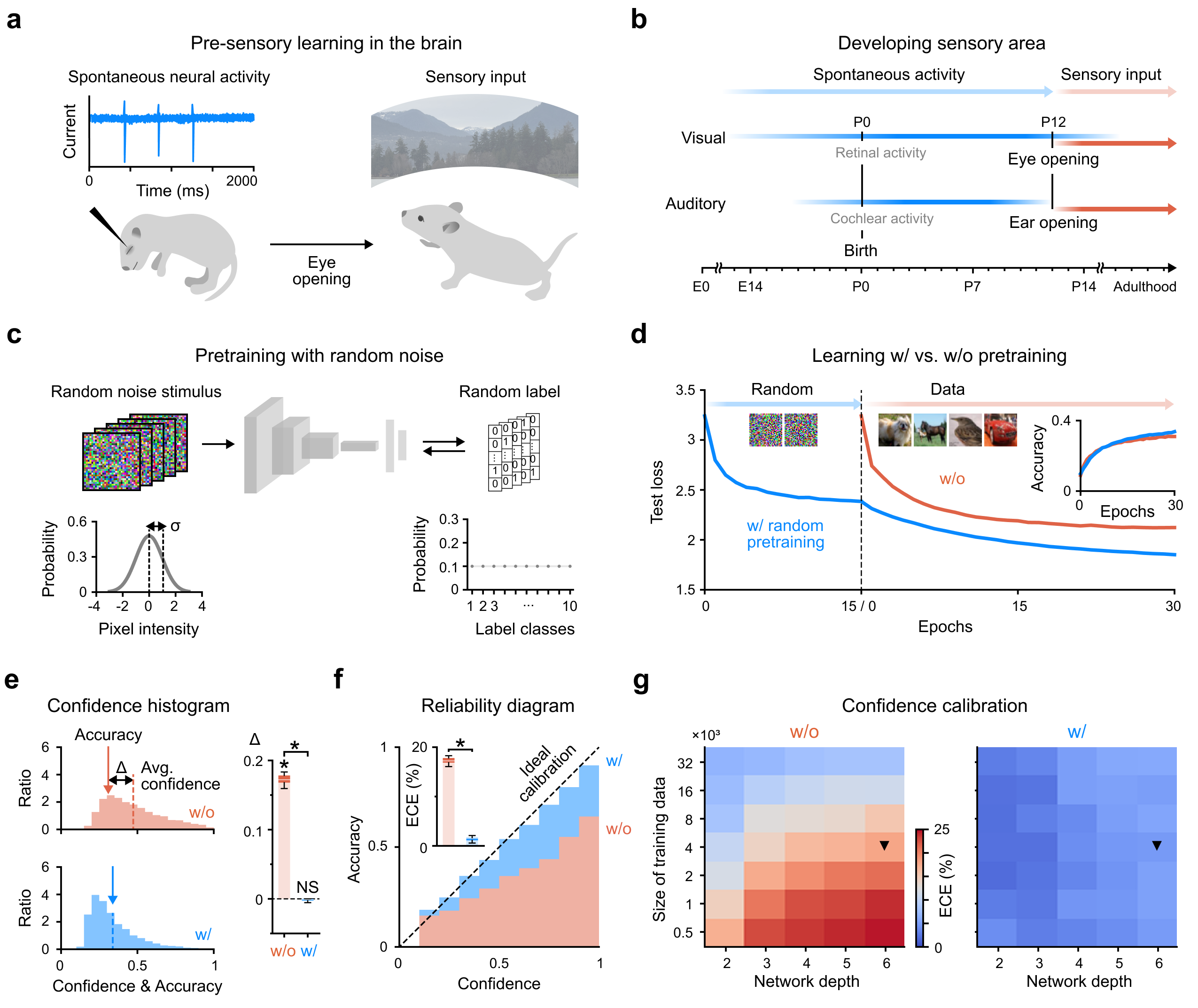}
    \caption{
        \textbf{Pretraining with random noise enables confidence calibration in neural networks.}\\
        (a) Prenatal learning through spontaneous neural activity prior to sensory input in a fetal rat, before birth (left) and after eye opening (right) \cite{galli1988}.
        (b) Spontaneous neural activity in the developing visual and auditory areas \cite{martini2021}.
        (c) Schematic of the pretraining algorithm with random noise, inspired by the developing brain before sensory experience. During pretraining, the network is trained on randomly sampled inputs from a Gaussian distribution and unpaired labels from a uniform distribution.
        (d) Test loss during random noise pretraining and data training. (Inset) Accuracy during data training in networks with (blue) and without (orange) pretraining.
        (e) Histogram of confidence for a network trained only with data (orange) and a network pretrained with random noise (blue). Averaged accuracy and confidence are indicated by vertical lines. (Right) The difference between averaged confidence and accuracy of predictions.
        (f) Reliability diagram showing the expected accuracy for samples binned by confidence. (Inset) Expected Calibration Error.
        (g) The effect of random noise pretraining under varying conditions of training data size and network complexity on calibration error. The marker indicates the parameters used in (d-f), where the six-layer networks are trained with a dataset size of 4,000.
    }
    \label{fig2}
\end{figure}

The biological brain begins its learning process even before birth, prior to encountering sensory inputs, through spontaneous neural activity \cite{galli1988} (Fig. \ref{fig2}a). During this early stage, the brain optimizes its neural connections by transmitting this spontaneous activity to higher cortical regions, and this prenatal learning is observed across multiple brain areas involved in sensory systems \cite{ackman2012, anton2019, galli1988, martini2021} (Fig. \ref{fig2}b). In particular, the spontaneous activities take on various spatiotemporal forms, such as waves \cite{ackman2012} or local synchronization \cite{anton2019}. However, when considered as a whole, they exhibit statistically random patterns. Thus, this process can be seen as a form of pretraining with random noise \cite{cheon2024}. Here, we demonstrate that a pretraining algorithm inspired by this biological process facilitates the preconditioning of networks, calibrating their confidence to a chance level.

We pretrained neural networks using noise inputs randomly sampled from a Gaussian distribution and random labels sampled from a uniform distribution (Fig. \ref{fig2}c). During training with random noise, we observed that the loss gradually decreased, while accuracy remained at the chance level (Fig. \ref{fig2}d; Supplementary Fig. \ref{figs2}). When the network was subsequently trained with real data after the noise pretraining, we observed a more substantial reduction in test loss, which was significantly lower compared to networks trained without random noise pretraining. Additionally, the network pretrained with random noise achieved higher accuracy (Fig. \ref{fig2}d, Inset; Supplementary Fig. \ref{figs2}). These results demonstrate that pretraining with random noise facilitates faster and more effective loss reduction during subsequent training with real data.

Next, we investigated whether random noise pretraining could improve the calibration of probabilities, aligning confidence with accuracy. We began by measuring the confidence distribution of the network’s output on the test dataset and compared it to the actual accuracy (Fig. \ref{fig2}e). We found that random noise pretraining significantly reduced the difference between confidence and accuracy during subsequent data training (Fig. \ref{fig2}e, Right, w/o vs. w/ random pretraining, $n_{\text{net}} = 10$, Wilcoxon rank-sum test, $P < 10^{-3}$; w/o vs. zero, Wilcoxon signed-rank test, NS, $P < 10^{-3}$; w/ vs. zero, Wilcoxon signed-rank test, NS, $P = 0.432$). We further examined this difference using a reliability diagram (Fig. \ref{fig2}f), which showed that the network pretrained with random noise had outputs that were closer to ideal calibration than networks trained on data alone. Additionally, we confirmed that the pretrained networks exhibited significantly lower calibration error compared to those trained only with data (Fig. \ref{fig2}f, Inset, w/o vs. w/ random pretraining, $n_{\text{net}} = 10$, Wilcoxon rank-sum test, $P < 10^{-3}$). We then extended this analysis to networks with varying depths and sizes of training data. We found that random noise pretraining had a significant effect, regardless of network depth or training data size (Fig. \ref{fig2}g, w/o vs. w/ random pretraining, $n_{\text{net}} = 10$, Wilcoxon rank-sum test, $P < 10^{-3}$). Pretraining successfully reduced calibration error across all conditions examined. Notably, the effect was particularly pronounced for networks with deeper structures trained on smaller datasets, where confidence miscalibration is more common. In addition to the results obtained using the standard backpropagation \cite{rumelhart1986} training method, we also examined networks trained with a biologically plausible algorithm employing feedback alignment \cite{lillicrap2016} (Supplementary Fig. \ref{figs3}). We observed that miscalibration also occurred in this model, but random noise pretraining significantly improved confidence calibration. This finding suggests that random noise pretraining could serve as a general solution for enhancing confidence calibration across various network learning models.

\subsection{Pre-calibration of network uncertainty across the input space using random noise}

\begin{figure}[t!]
    \centering
    \includegraphics[width=\textwidth]{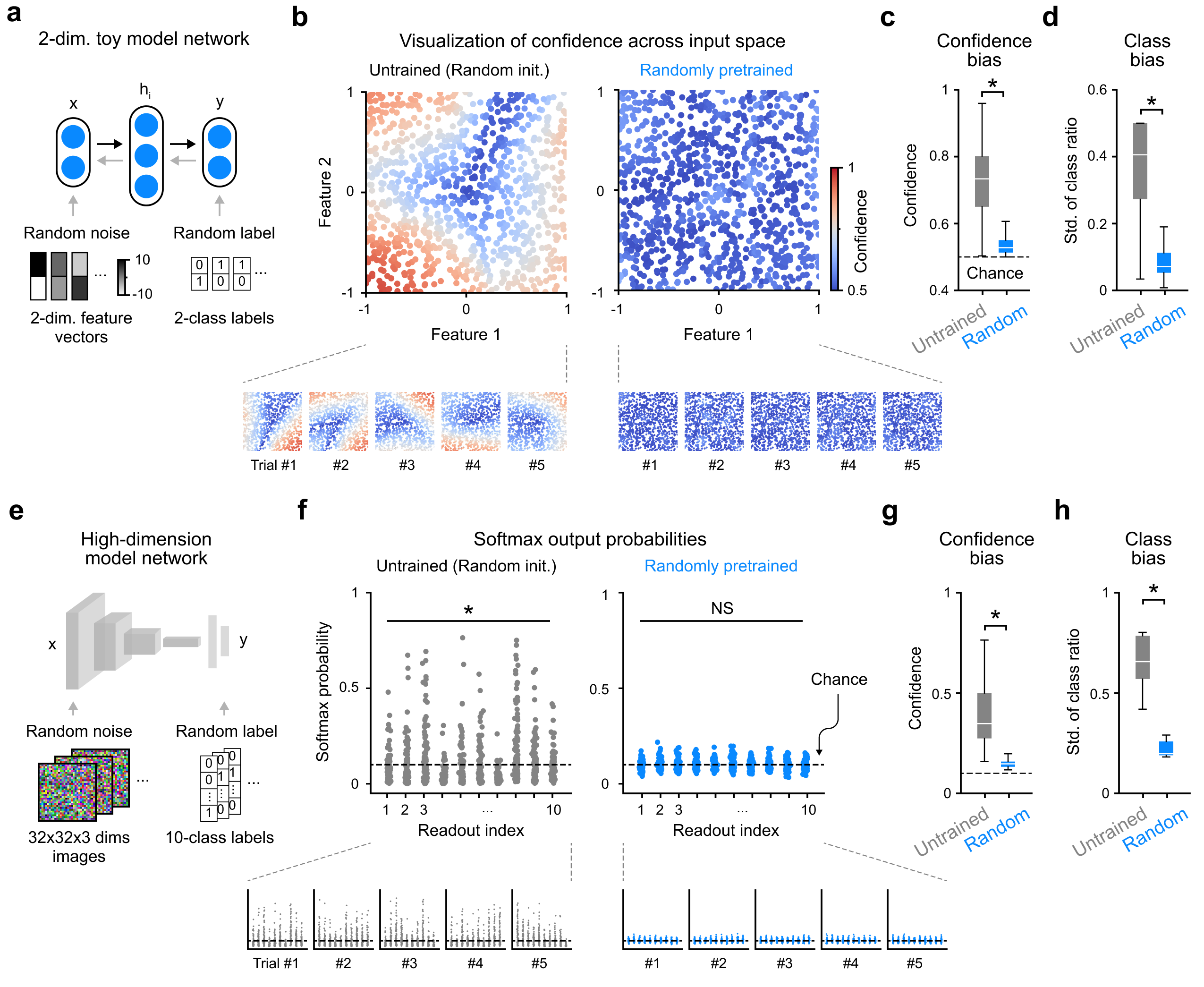}
    \caption{
        \textbf{Random noise pre-calibrates neural network uncertainty over input space.}\\
        (a) Diagram illustrating the training of random noise in a toy-model network with a two-dimensional input space and binary output space. The network receives random noise with two input features and randomly assigned binary labels, and is trained using Binary Cross-Entropy (BCE) loss.
        (b) Visualization of confidence over the input space. (Left) Confidence map of the untrained network. (Right) Confidence map after random noise pretraining.
        (c) Confidence distribution of the network for two-dimensional noise.
        (d) Class bias of the network for two-dimensional noise, showing the extent of bias in class prediction.
        (e) Diagram of training random noise in a multi-layer perceptron with a high-dimensional input space, designed to classify the CIFAR-10 dataset. The network receives 32×32×3 input images and outputs SoftMax probabilities for ten classes.
        (f) Visualization of SoftMax output probabilities for random noise inputs in the readout neurons. (Left) Untrained network. (Right) Network after random noise pretraining. The dashed line represents the chance level of classification.
        (g) Confidence distribution of the network.
        (h) Class bias of the network, showing the bias towards specific classes.
    }
    \label{fig3}
\end{figure}

To better understand how pretraining with random noise affects confidence calibration, we further explored the characteristics of randomly initialized, untrained networks and those that were randomly pretrained, prior to exposure to real data. For this purpose, we introduced a simplified toy-model network with a two-dimensional input space (Fig. \ref{fig3}a), allowing for visualization of the input feature space (Fig. \ref{fig3}b). Under these conditions, we observed that the initial state of a conventionally processed network — randomly initialized \cite{he2015} and untrained — displayed a highly biased distribution of confidence across the input space (Fig. \ref{fig3}b, Left). Moreover, we confirmed that this initial miscalibration was consistent across multiple trials with randomly initialized networks (Supplementary Fig. \ref{figs4}a). In contrast, networks pretrained with random noise displayed a more homogeneous confidence distribution across the input space, effectively calibrating to the chance level (Fig. \ref{fig3}b, Right; Supplementary Fig. \ref{figs4}b). This result demonstrates that pretraining with random noise helps mitigate overconfidence by calibrating confidence levels across the entire input space, bringing them closer to chance (Fig. \ref{fig3}c, w/ vs. w/o random pretraining, $n_{\text{trial}} = 1000$, Wilcoxon rank-sum test, $P < 10^{-3}$; Supplementary Fig. \ref{figs4}c). We also observed that untrained networks could exhibit initial biases toward specific output classes due to fluctuations in confidence. However, random noise pretraining significantly reduced this initial class bias, leading to a more uniform confidence distribution across the classes (Fig. \ref{fig3}d, w/ vs. w/o random pretraining, $n_{\text{net}} = 10$, Wilcoxon rank-sum test, $P < 10^{-3}$).

Next, we extended this analysis to models with higher-dimensional input spaces (Fig. \ref{fig3}e). Specifically, we used a model with input features of 32×32×3 dimensions and output SoftMax probabilities for ten classes, allowing it to train on CIFAR-10 data. We first measured the probability outputs for random noise inputs in both untrained and randomly pretrained networks (Fig. \ref{fig3}f; Supplementary Fig. \ref{figs5}). Consistent with the toy model results, we observed overconfidence and class bias in untrained networks (Fig. \ref{fig3}f, Left, $n_{\text{trial}} = 100$, one-way ANOVA, $P < 10^{-3}$; Supplementary Fig. \ref{figs5}a), whereas such biases were barely detectable in randomly pretrained networks (Fig. \ref{fig3}f, Right, $n_{\text{trial}} = 100$, one-way ANOVA, NS, $P = 0.429$; Supplementary Fig. \ref{figs5}b). We confirmed that these differences aligned with the toy model results, both in terms of confidence distribution (Fig. \ref{fig3}g, w/ vs. w/o random pretraining, $n_{\text{trial}} = 100$, Wilcoxon rank-sum test, $P < 10^{-3}$; Supplementary Fig. \ref{figs5}c) and class bias (Fig. \ref{fig3}h, w/ vs. w/o random pretraining, $n_{\text{net}} = 10$, Wilcoxon rank-sum test, $P < 10^{-3}$). These results suggest that random noise pretraining enables the network to maintain lower, more uniformly calibrated confidence levels, even when handling more complex, higher-dimensional input data.

We further investigated whether the pre-calibration effect observed with random noise training extends to unseen, more realistic data. Specifically, we examined how the network's confidence on the CIFAR-10 and SVHN \cite{netzer2011} datasets changes during random noise pretraining (Supplementary Fig. \ref{figs6}a, Left) by measuring the network's predictions for real data after each epoch of pretraining. Notably, we observed that the confidence for unseen CIFAR-10 data was properly calibrated throughout the random noise training phase, approaching the chance level, while accuracy remained at chance level (Supplementary Fig. \ref{figs6}a, Right). We confirmed that this trend was consistent across different unseen datasets, including CIFAR-10 (Supplementary Fig. \ref{figs6}b, CIFAR-10; w/ vs. w/o random pretraining, $n_{\text{trial}} = 10000$, Wilcoxon rank-sum test, $P < 10^{-3}$) and SVHN (Supplementary Fig. \ref{figs6}b, SVHN; w/ vs. w/o random pretraining, $n_{\text{trial}} = 10000$, rank-sum test, $P < 10^{-3}$). The network maximized uncertainty for unseen data, with accuracy remaining at the chance level, thereby aligning confidence with accuracy. As a result, calibration error — the discrepancy between confidence and accuracy — was significantly reduced for unseen data (Supplementary Fig. \ref{figs6}c, w/ vs. w/o random pretraining, $n_{\text{net}} = 10$, Wilcoxon rank-sum test, CIFAR-10, $P < 10^{-3}$; SVHN, $P < 10^{-3}$). These findings demonstrate that conventionally initialized networks tend to exhibit overconfidence and class bias even before training on data, but random noise pretraining effectively eliminates this initial bias and calibrates confidence to the chance level, maximizing uncertainty for unseen datasets. This suggests that random noise pretraining can serve as a novel initialization strategy, offering an alternative to conventional random initialization(He et al., 2015).

\subsection{Aligning confidence levels with actual accuracy}

\begin{figure}[t!]
    \centering
    \includegraphics[width=\textwidth]{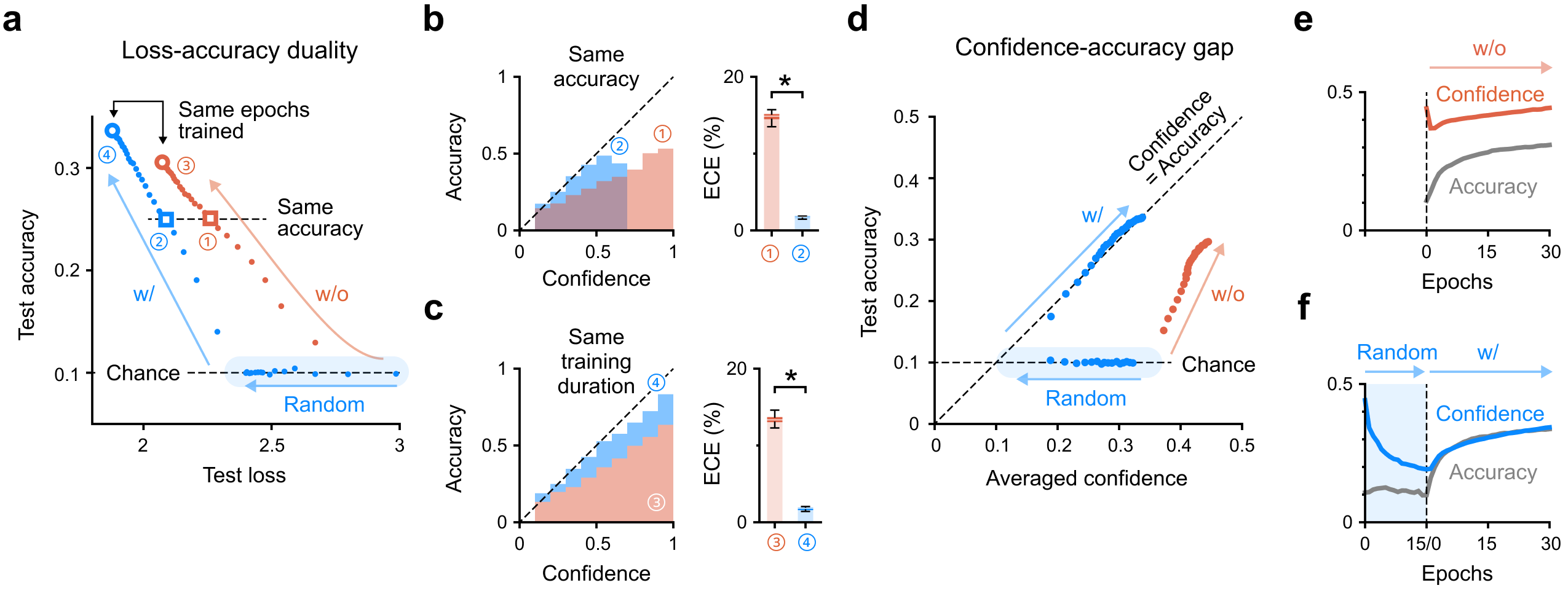}
    \caption{
        \textbf{Pre-calibration enables learning with matching confidence and accuracy.} \\
        (a) Visualization of loss-accuracy duality. Each dot represents the test loss and accuracy at each epoch of training. Four points are selected for calibration comparison: networks achieving the same accuracy for both the random noise-pretrained network and the data-only training network (\textcircled{1}, \textcircled{2}), as well as fully converged networks trained for the same number of epochs (\textcircled{3}, \textcircled{4}).
        (b-c) Comparison between random noise-pretrained networks and data-only training networks. (Left) Reliability diagram, showing the relationship between predicted confidence and accuracy. (Right) Calibration error, measured as Expected Calibration Error (ECE) (b) Accuracy-controlled comparison (\textcircled{1} vs. \textcircled{2}).
        (c) Epoch-controlled comparison (\textcircled{3} vs. \textcircled{4}).
        (d) Scatter plot of confidence versus accuracy for the training dataset in a data-only training network (orange) and a random noise pretrained network (blue). The diagonal line indicates perfect calibration, where confidence matches expected accuracy.
        (e-f) Averaged confidence and accuracy measured across training epochs for the training dataset.
        (e) Data-only training network.
        (f) Random noise-pretrained network.
    }
    \label{fig4}
\end{figure}

To better understand the impact of pre-calibration, we examined the learning dynamics of networks during data training, by analyzing the training trajectory in a two-dimensional plane of accuracy and loss (Fig. \ref{fig4}a). The trajectory showed significant differences between networks with and without random noise pretraining, primarily due to the initial loss reduction during pretraining. Specifically, for points on both curves with the same accuracy, the randomly trained network exhibited significantly lower loss than the network trained solely on data. By selecting arbitrary points of identical accuracy (Fig. \ref{fig4}a, \textcircled{1} and \textcircled{2}), we analyzed the reliability diagram and calibration error for both networks. We found that networks with random pretraining had better confidence calibration compared to those without pretraining, even with the same accuracy (Fig. \ref{fig4}b, w/o vs. w/ random pretraining, $n_{\text{net}} = 10$, Wilcoxon rank-sum test, $P < 10^{-3}$). Similarly, when selecting arbitrary points with the same total training epochs (Fig. \ref{fig4}a, \textcircled{3}, \textcircled{4}), pretrained networks again demonstrated better confidence calibration (Fig. \ref{fig4}c, w/o vs. w/ random pretraining, $n_{\text{net}} = 10$, Wilcoxon rank-sum test, $P < 10^{-3}$). These results highlight how random noise pretraining influences the dynamics of subsequent learning, particularly by reducing initial loss.

We next explored the relationship between confidence and accuracy throughout the entire training process (Fig. \ref{fig4}d). In pretrained networks, confidence and accuracy remained well-aligned, consistently following the diagonal line during training (Fig. \ref{fig4}d, w/). In contrast, networks without pretraining did not exhibit this alignment (Fig. \ref{fig4}d, w/o and Fig. \ref{fig4}e). Notably, a significant disparity between confidence and accuracy was initially observed due to early miscalibration, but this gap quickly diminished and disappeared during random noise pretraining (Fig. \ref{fig4}d, Random). This alignment of confidence and accuracy supports more synchronized learning during subsequent data training (Fig. \ref{fig4}f). Thus, calibrating confidence to the chance level through random noise pretraining ensures proper confidence calibration.

\subsection{Detection of out-of-distribution samples using calibrated network confidence}

\begin{figure}[t!]
    \centering
    \includegraphics[width=120mm]{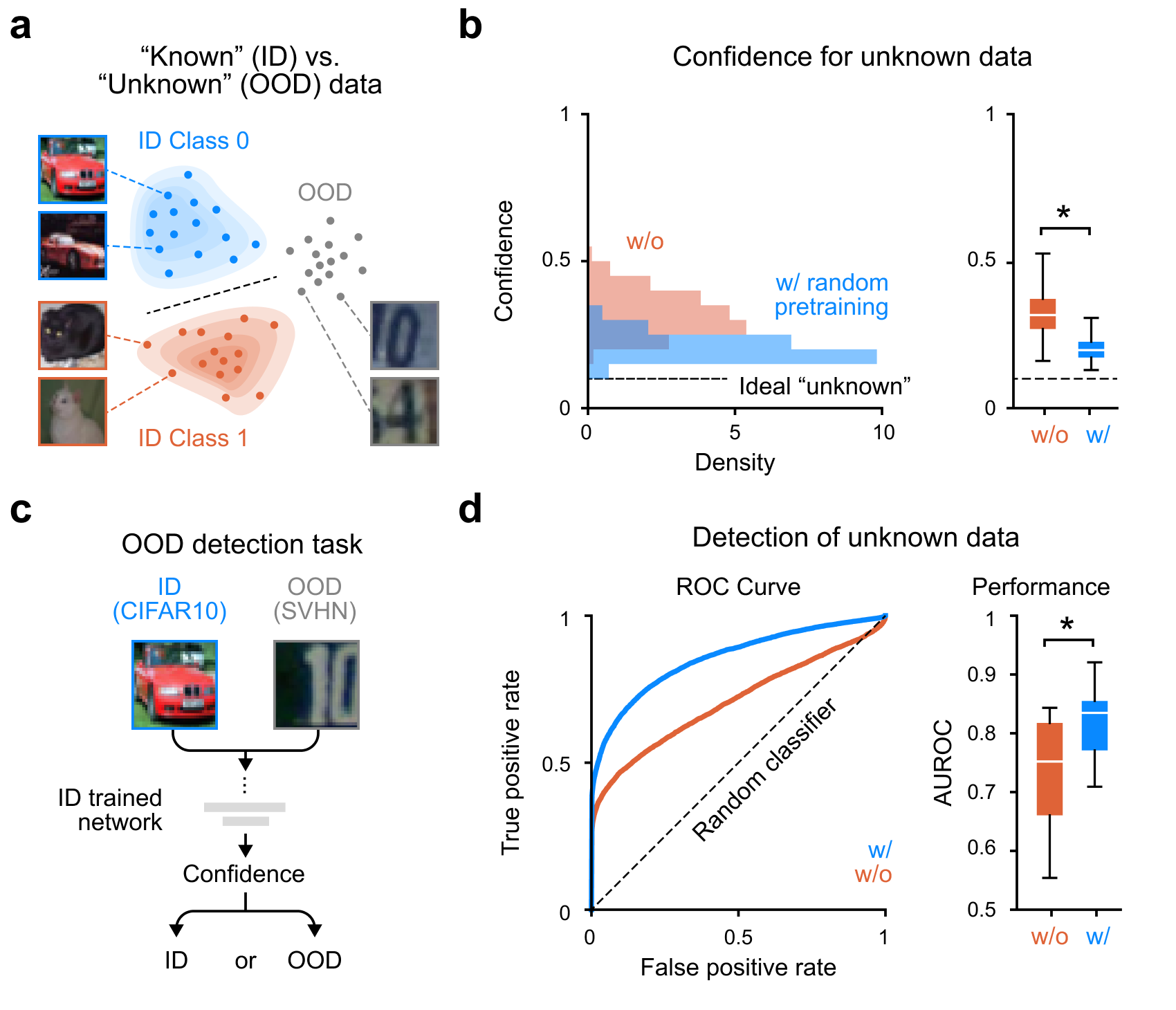}
    \caption{
        \textbf{Out-of-distribution detection using calibrated network confidence.}\\
        (a) Conceptual illustration of In-Distribution (ID) and Out-Of-Distribution (OOD) samples. ID samples are those that the model has been trained on, while OOD samples are from a different distribution, not seen during training. CIFAR-10 is used as ID data to train the neural network, and SVHN is used as OOD data, which the network has not been trained on.
        (b) Histogram of confidence for unseen OOD data (left). The distribution is also displayed as a boxplot (right). The horizontal line represents the chance level, indicating ideally calibrated confidence for unknown data, where the model exhibits maximized uncertainty.
        (c) Schematic of the OOD detection task. After training the network on ID data, both ID and OOD samples are presented to the network, and confidence is measured to determine whether the samples belong to the ID or OOD category.
        (d) Performance of the OOD detection task using network confidence. (Left) Receiver Operating Characteristic (ROC) curve, where the diagonal line represents random classification, and the point (0, 1) represents an ideal classifier. (Right) Performance of the OOD detection task, measured by the area under the ROC curve (AUROC).
    }
    \label{fig5}
\end{figure}

Finally, we examined the effect of random noise pretraining on detecting out-of-distribution (OOD) data. OOD samples refer to “unseen” data the model has not been trained on, while in-distribution (ID) samples are those the model has been trained to recognize (Fig. \ref{fig5}a). After training on CIFAR-10 (ID), we evaluated the model's response to unseen SVHN data (OOD). Networks without random noise pretraining exhibited high confidence on OOD samples, even though they had no prior knowledge of them (Fig. \ref{fig5}b, Left). In contrast, networks with random pretraining showed significantly lower confidence on OOD samples, aligning closer to the chance level — ideal for “unknown” data. We confirmed that randomly pretrained networks had lower average confidence for OOD samples, closer to chance level (Fig. \ref{fig5}b, Right; w/ vs. w/o random pretraining, $n_{\text{sample}} = 10000$, Wilcoxon rank-sum test, $P < 10^{-3}$).

We further explored how calibrated confidence for both in-distribution (ID) and out-of-distribution (OOD) samples can improve OOD detection. To achieve this, we implemented a simple framework where raw confidence values were used to distinguish ID from OOD samples (Fig. \ref{fig5}c), with samples below an arbitrary threshold being classified as OOD. We then assessed the OOD detection performance across various thresholds and used these results to plot the receiver operating characteristic (ROC) curve (Fig. \ref{fig5}d, Left). To quantify performance, we calculated the area under the ROC curve (AUROC) (Fig. \ref{fig5}d, Right) and found that random noise pretraining significantly improved OOD detection (Fig. \ref{fig5}d, Right; w/ vs. w/o random pretraining, $n_{\text{net}} = 10$, Wilcoxon rank-sum test, $P < 10^{-3}$). These results suggest that calibrated confidence maintains low confidence for unseen OOD data, enabling robust discrimination between “known” and “unknown” data without the need for post-processing or additional computations.

\section{Discussion}

The ability to estimate uncertainty and recognize the unknown is crucial for real-world intelligent systems. Human intelligence naturally exhibits meta-cognition — the ability to assess probabilistic confidence \cite{cosmides1996, masset2020} and distinguish between what is known and unknown. This capacity plays a crucial role in complex human behaviors, particularly in decision-making \cite{fleming2012, kepecs2008, kiani2009}. Without this meta-cognitive ability to estimate uncertainty, intelligent systems may make critical errors in real-world scenarios where incorrect decisions can be costly \cite{helldin2013, mehrtash2020}. Despite recent advances in performance \cite{lecun1998, Krizhevsky2012, he2016, huang2017}, deep neural networks continue to struggle with uncertainty calibration \cite{guo2017, nalisnick2019, ovadia2019}, as our findings have demonstrated. This issue poses a significant challenge for integrating artificial intelligence into daily life, despite its advanced capabilities.

Hallucinations in large language models (LLMs) \cite{achiam2023, team2023} are a prominent example of how miscalibrated uncertainty can make artificial intelligence systems unreliable. While LLMs exhibit expert-level proficiency across various domains, such as science, programming, and healthcare, they often produce incorrect answers with high confidence, making it difficult to trust their outputs in critical situations \cite{farquhar2024, xiao2021}. Beyond LLMs, miscalibration has long been a challenge in applying AI across real-world domains. For instance, although autonomous vehicle prototypes emerged over a decade ago, their integration into daily life remains difficult due to reliability concerns, highlighted by a fatal accident caused by the control system misinterpreting an object \cite{vlasic2016}. Similar issues arise in healthcare, where models that classify diseases from radiological images often make overconfident but incorrect predictions when presented with outlier inputs \cite{Wollek2024}. Despite achieving expert-level accuracy on medical board exams, LLMs frequently fail to recognize their knowledge limitations and provide incorrect yet confident answers \cite{griot2025}. Relying on such overconfident judgments in medical procedures could jeopardize patient safety. These examples highlight how miscalibrated predictions can lead to catastrophic decision-making errors, emphasizing the need to address miscalibration issues before AI is deployed in critical applications.

Our results demonstrate that these miscalibrations of uncertainty are inevitable in modern deep neural networks as they become more complex to enhance their learning capacity \cite{lecun1998, Krizhevsky2012, he2016, huang2017}. We found that both dataset size and model complexity systematically influence calibration quality — when the dataset is insufficient, large and complex models experience significant calibration errors. This raises the question of whether an extremely large dataset is always necessary for achieving optimal calibration as model complexity increases. However, acquiring high-quality training data is often costly in the real world, and in some domains, the available data is inherently limited, such as text data in minority languages or medical data on rare diseases. Given these practical challenges, relying solely on expanding dataset size is not a feasible solution as model complexity grows. Our proposed method addresses this by decoupling miscalibration issues from both dataset size and model complexity, offering a simple yet effective strategy for modern deep neural networks as they evolve with increasing parameters.

A longstanding question in machine learning is how to properly initialize neural networks for efficient learning, as the network may converge to a local minimum or diverge depending on the initial condition \cite{glorot2010, sutskever2013}. Previous studies have shown that adjusting initialization can help networks converge with higher accuracy \cite{he2015, saxe2013}, leading to the development of optimized techniques for initialization, a common starting point for training. However, our findings suggest that these traditional methods may not be ideal for learning due to issues with confidence calibration — networks initialized by conventional methods tend to suffer from overconfidence and biases toward specific outputs. The widely adopted practice of random initialization in deep learning is actually a key factor contributing to the miscalibration of network uncertainty; however, pretraining with random noise offers a straightforward and effective solution, thereby proposing a new approach for network initialization. This approach is applicable to all types of network models, and remains effective regardless of model size, complexity, or other characteristics. Unlike previous uncertainty calibration methods that rely on auxiliary computation or post-processing \cite{guo2017, hendrycks2018, lee2018b, liu2020, niculescu2005, platt1999, zadrozny2001, zadrozny2002}, our findings demonstrate that simply altering the initialization of learning through random noise pretraining significantly changes learning dynamics, aligning confidence with accuracy.

Our work on pretraining with random noise is inspired by the prenatal developmental process of biological brains. While spontaneous neural activity in the brain \cite{ackman2012, anton2019, galli1988, martini2021} is recognized as important for development, its functional role remains unclear. Recently, artificial neural networks have been considered as a powerful framework for understanding biological neural processing \cite{hassabis2017, richards2019, yamins2016}. Our results not only provide an effective and practical solution to calibration issues in machine learning models, but also offer insights into biological prenatal processes. In addition to employing the backpropagation algorithm \cite{rumelhart1986}, we explored the benefits of random noise pretraining within biologically plausible learning rules, such as feedback alignment \cite{lillicrap2016}. Interestingly, we observed severe miscalibration when using local learning rules, but random noise pretraining improved calibration substantially, consistent with our primary findings. Our recent research also indicated that initial calibration and conditioning through random noise pretraining can significantly enhance the learning efficiency of biologically plausible learning systems \cite{cheon2024}. These findings suggest that the pre-sensory learning process in the biological brain may involve learning the “unknown” before acquiring knowledge through sensory experience, potentially marking a key developmental step toward meta-cognition.

While the current study provides valuable insights, it is important to acknowledge its limitations. Our model was not directly validated through real-world applications or benchmarking, and the findings are based on relatively simple tasks and smaller-scale networks compared to state-of-the-art models, such as large language models (LLMs). Given that training a single LLM can incur millions of dollars in computational costs, it is often impractical to systematically investigate their calibration in a typical laboratory setting. However, we are confident that the results obtained in this study can be replicated in more complex, real-world models, whether in image classification, text generation, or other domains. A key characteristic across various architectures and task domains is that the final outputs are probability distributions. Our pretraining method remains broadly applicable and effective as long as the model generates probability distributions for final decision-making, regardless of its structure, scale, or domain. Moving forward, applying our model to a range of commercial AI systems to further validate and refine these results presents a valuable avenue for future research.

In summary, we conclude that pretraining with random noise effectively pre-calibrates network uncertainty, enabling subsequent learning with aligned accuracy and confidence. These results highlight the importance of the initial state of neural networks for effective data learning \cite{kim2021, baek2021, cheon2022, lee2023, glorot2010, he2015}, as well as the significance of the early developmental process before encountering data \cite{kim2020, cheon2024}. Our findings provide new insights into uncertainty calibration, offering potential solutions to challenges arising from calibration issues, such as machine hallucination.

\newpage
\section{Methods}

\subsection*{Neural network model}

We employed a multi-layer feedforward model as a simplified representative of conventional neural networks to investigate the miscalibration problem and the effect of random noise pretraining. The model consists of multiple non-linear layers, each performing a series of operations: multiplying weights, adding biases, and passing through a ReLU activation function. To mitigate overfitting, batch normalization was applied to every layer. The hidden layer size was fixed at 256 neurons, with the layer depth varied across different experiments. The network weights were randomly initialized following a Gaussian distribution, with a mean of 0 and a standard deviation determined to control the gain across layers, as per the standard initialization method \cite{he2015}. The biases were initialized to zero. We observed that our results were fairly consistent regardless of hyperparameter settings, such as the hidden layer size, under typical conditions.

\subsection*{Pretraining with random noise}

The random noise input was sampled from a Gaussian distribution with a mean of 0 and a standard deviation of 1. We set the noise size to 32×32×3, matching the size of the input data and the input dimension of the neural networks. The corresponding labels for the random noise were also randomly sampled from a uniform distribution and one-hot encoded. Both the random noise inputs and their labels were re-sampled at each iteration and were not paired with one another. During training, the random noise inputs were fed into the neural network, and the model output was obtained. The error was calculated as the distance between the model output and the random labels, and the network was trained to minimize this error, similar to conventional training procedures.

\subsection*{Subsequent training with real data}

After pretraining the neural network with random noise, we proceeded to train it with real data. We used a subset of the CIFAR-10 \cite{krizhevsky2009} dataset, which consists of 32×32×3 natural images representing 10 categories: airplanes, automobiles, birds, cats, deer, dogs, frogs, horses, ships, and trucks. In the primary analysis, we used a training subset of 1000 samples to assess the robustness of calibration following random noise pretraining under challenging conditions. To investigate the effects of training data size and model capacity, we employed feedforward networks with depths of 2, 3, 4, 5, and 6 and trained them on CIFAR-10 subsets of sizes 500, 1000, 2000, 4000, 8000, 16000, and 32000.

For both the random noise pretraining and subsequent training with real data, we followed commonly used neural network training procedures. We applied standard backpropagation with the Adam optimizer, using learning rates between 0.0001 and 0.00001 and betas of (0.99, 0.999). To prevent overfitting, we used weight decay with a decay constant of 0.001. During the analysis, we compared the networks pretrained with random noise to those trained solely on data, without any random noise pretraining. All conditions, except for the random noise pretraining, were carefully controlled to isolate the effect of the random noise pretraining. 

Additionally, we analyzed the neural network training using a feedback alignment \cite{lillicrap2016} algorithm, as a biologically plausible alternative to backpropagation that does not require weight transport. In feedback alignment, random synaptic feedback is used to compute the error signal, instead of using forward weights. Results obtained using this method are presented in the supplementary material.

\subsection*{Evaluating calibration of neural networks}

To evaluate the calibration of the model, we used a reliability diagram \cite{degroot1983, niculescu2005}, which visualizes the relationship between the model's confidence and its actual accuracy. In this approach, we bin the model's confidence and compute the expected accuracy within each bin. The reliability diagram then plots the expected accuracy as a function of the model's confidence
\begin{equation}
\text{acc}(B_m) = \frac{1}{B_m} \sum_{i=1}^{B_m} I (\text{pred}_i = \text{label}_i)
\end{equation}
\begin{equation}
\text{conf}(B_m) = \frac{1}{B_m} \sum_{i=1}^{B_m} \text{conf}_i
\end{equation}
where $M$ is the total number of bins, $B_m$ is the number of samples in the $m$-th bin, $N$ is the total number of samples, $\text{acc}(B_m)$ is the accuracy of the $m$-th bin, and $\text{conf}(B_m)$ is the average confidence in the $m$-th bin. In our simulations, we used 10 bins.

To quantitatively measure the calibration of the model, we used the Expected Calibration Error (ECE) \cite{naeini2015}, which calculates the average difference between the model’s confidence and its actual accuracy within the binned confidence. ECE is calculated as: 
\begin{equation}
\text{ECE} = \sum_{m=1}^{M} \frac{B_m}{N} \left| \text{acc}(B_m) - \text{conf}(B_m) \right|
\end{equation}

In an ideally calibrated model, the expected accuracy perfectly matches its confidence. Practically, a model is considered well-calibrated if its ECE is close to zero.

\subsection*{Visualization of confidence in toy-model and high-dimensional networks}

As a proof-of-concept, we first employed a simple toy-model network with a two-dimensional (2D) input space. The input space is defined as the span of two features, each varying between -1 and 1. This allows us to visualize the input space in the 2D plane, where the x-axis represents the first feature, and the y-axis represents the second feature. The neural network model consists of two layers, with a hidden layer of size 10. It outputs a two-dimensional vector for binary classification. The model is initialized using He initialization. 

We trained this toy-model network with 2-dimensional random noise. Subsequently, we measured the model’s confidence for various inputs and plotted the resulting confidence on the input space, comparing both untrained and random-pretrained networks. This enables the visualization of the model's confidence map across the input space. In addition, we analyzed the class bias of the model. The model outputs both the predicted class and its associated confidence. We calculated the ratio of predictions for each class and computed the standard deviation of these ratios:
\begin{equation}
 \text{Class bias} = \text{std}([N_0, N_1, ..., N_M]/\sum_{i=1}^{M}N_i)
    \end{equation}
where ${N_i}$ represents the count of outputs predicted to class $i$, and $M$ is the total number of output classes. A high class bias indicates that the model is biased toward specific readout neurons, while a class bias close to zero suggests that the model is unbiased, with predictions distributed more uniformly across the readout neurons.

To extend this analysis, we performed similar experiments with high-dimensional feedforward neural networks. We sampled high-dimensional random noise and measured the SoftMax probability output of the network. We visualized the model’s SoftMax outputs for both untrained networks and those pretrained with random noise. During the random noise pretraining of the high-dimensional networks, we also measured the confidence and accuracy for real data, such as CIFAR-10 and SVHN \cite{netzer2011}. Additionally, we calculated the Expected Calibration Error (ECE) for CIFAR-10 and SVHN during random noise pretraining.

\subsection*{Out-of-distribution detection task}

To design the out-of-distribution (OOD) detection task, we used CIFAR-10 as the in-distribution (ID) dataset for training. For the OOD samples, we selected SVHN, which also has a 32×32×3 input size. After training the model on CIFAR-10, we measured the confidence of the network for both ID and OOD samples. Based on the measured confidence, we distinguished between ID and OOD samples by setting an arbitrary threshold. If the confidence exceeded the threshold, the sample was predicted as an ID sample; if the confidence was lower than the threshold, it was predicted as an OOD sample. Since the performance of OOD detection depends on the chosen threshold, we varied the threshold continuously from 0 to 1 and measured the corresponding false positive rate (FPR) and true positive rate (TPR). We then plotted the detection performance as a receiver operating characteristic (ROC) curve, with the true positive rate plotted against the false positive rate. To quantitatively assess the detection performance, we measured the area under the ROC curve (AUROC), which serves as a metric for the performance of OOD detection.

\section*{Statistical analysis}
All statistical variables, including sample sizes, exact P values, and statistical methods, are provided in the corresponding text or figure legends.

\section{Data availability}
The datasets used in this study are publicly available: \url{https://www.cs.toronto.edu/~kriz/cifar.html} (CIFAR-10), \url{http://ufldl.stanford.edu/housenumbers} (SVHN).

\section{Code availability}
The analysis and simulations were performed using Python 3.11 (Python software foundation) with PyTorch 2.1 and NumPy 1.26.0. Statistical tests were conducted using SciPy 1.11.4 was used to perform the statistical test and analysis. The code used in this work is available at \url{https://github.com/cogilab/Random2}.

\section{Acknowledgements}
This work was supported by the National Research Foundation of Korea (NRF) grants (NRF2022R1A2C3008991 to S.P.) and by the Singularity Professor Research Project of KAIST (to S.P.).

\section{Author contributions}
Conception and design: J.C. Development of the methodology: J.C. and S.P. Funding acquisition: S.P. Collection of data: J.C. Formal analyses: J.C. and S.P. Writing the original draft: J.C. and S.P. Writing, reviewing and/or revising the manuscript: J.C. and S.P.  Study supervision: S.P.

\section{Declaration of completing interests}
The authors declare that they have no competing interests.

\newpage
{
\small
\begin{spacing}{1.2}
\bibliographystyle{naturemag}
\bibliography{reference}
\end{spacing}
}

\clearpage
\section{Supplementary Materials}
\renewcommand{\figurename}{Supplementary Fig.}
\setcounter{figure}{0}

\begin{figure}[h!]
    \centering
    \includegraphics[width=120mm]{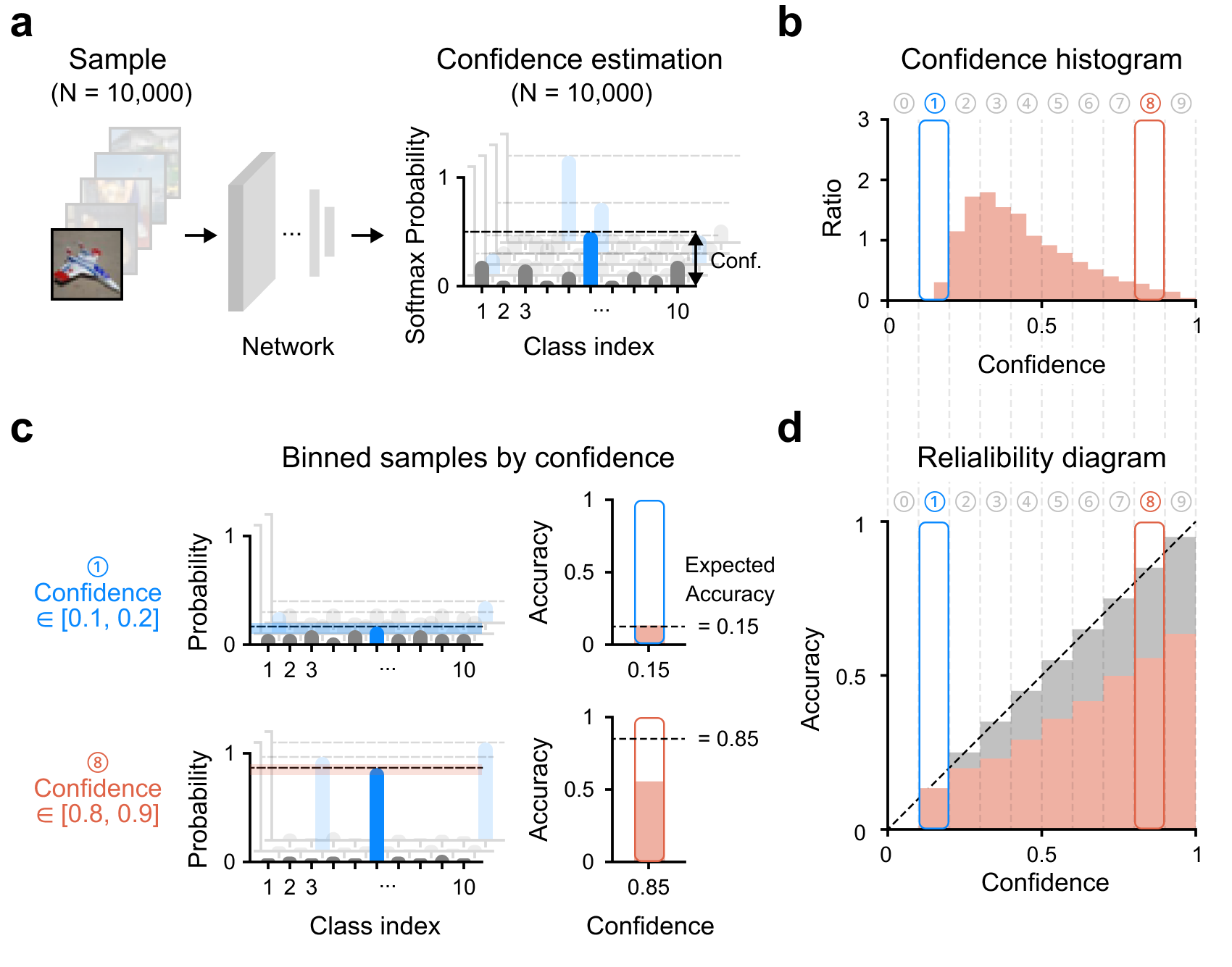}
    \caption{
        \textbf{Measuring confidence calibration of a model network through a reliability diagram.} \\
        (a) Sample images were presented to a trained neural network, and its predictions were recorded ($n_{\text{trial}} = 10000$). Specifically, confidence and predicted answers were measured in each trial.
        (b) Confidence histogram. Each trial is binned according to its confidence level.
        (c) In each batch of predictions binned by confidence, the correct ratios were measured. The diagram shows two example batches with prediction trials at different confidence levels. In an ideally calibrated model, the correct ratios in each batch should match the confidence levels.
        (d) Reliability diagram. This diagram shows the model’s accuracy as a function of confidence levels for each batch. In an ideally calibrated model (gray bars), accuracy should align with confidence levels across all predictions, represented by a diagonal line.
    }
    \label{figs1}
\end{figure}

\clearpage
\begin{figure}[t!]
    \centering
    \includegraphics[width=\textwidth]{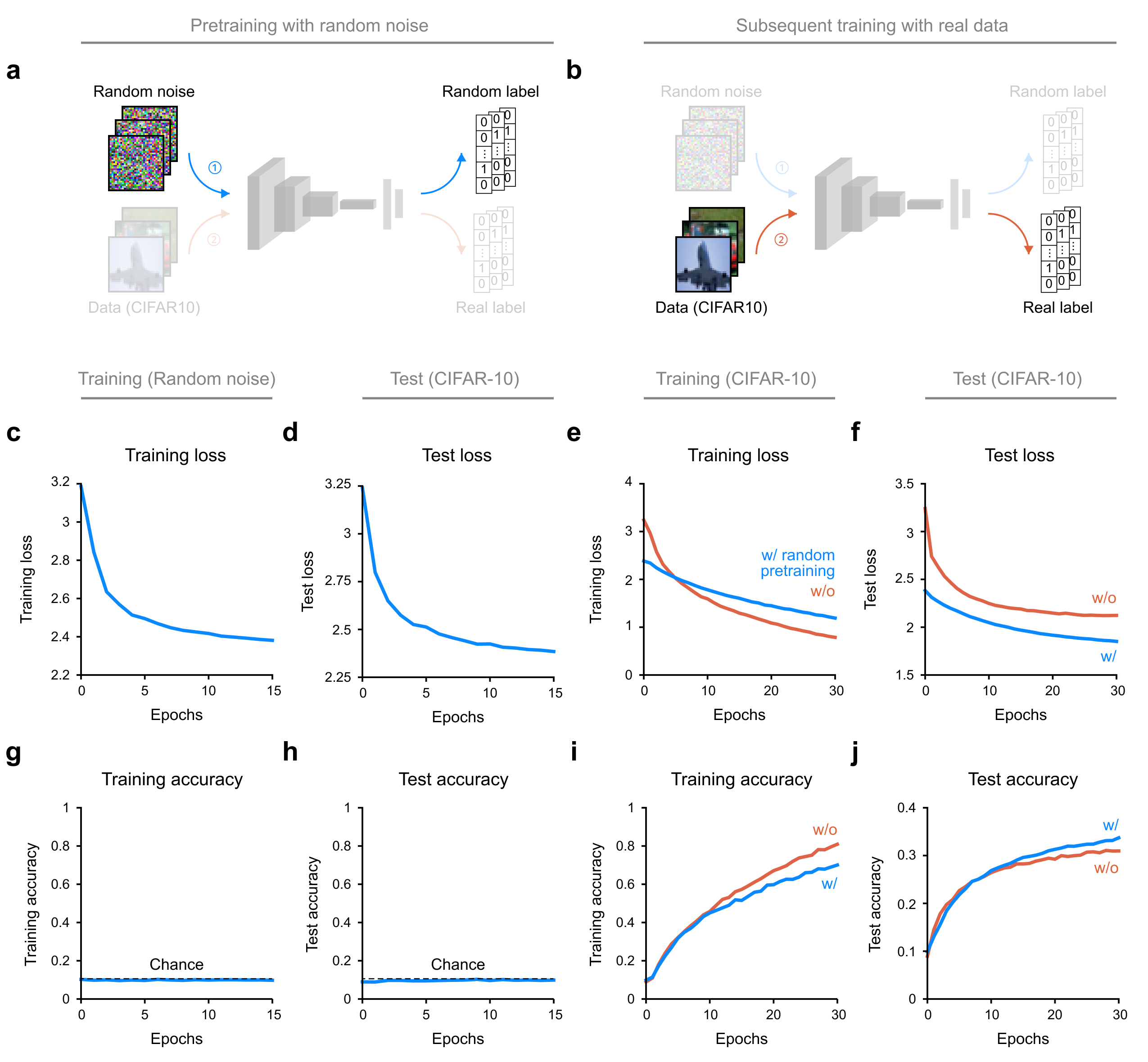}
    \caption{
        \textbf{Learning curve in random noise pretraining and subsequent data training stage.} \\
        (a) The pretraining process, where neural networks are trained on random noise before encountering real data.
        (b) After pretraining with random noise, the network is subsequently trained with real data.
        (c-d) Loss curves during random noise pretraining:
        (c) Training loss.
        (d) Test loss.
        (e-f) Loss curve during subsequent data training:
        (e) Training loss.
        (f) Test loss.
        (g-h) Accuracy curves during random noise pretraining:
        (g) Training accuracy.
        (h) Test accuracy. Note that the accuracy remains at chance level, as there is no explicit correlation between the input (random noise) and the output labels.
        (i-j) Accuracy curve during subsequent data training:
        (i) Training accuracy.
        (j) Test accuracy.
        In each learning curve, the orange line represents the network trained solely with data (without random noise pretraining), while the blue line represents the network pretrained with random noise.
    }
    \label{figs2}
\end{figure}

\clearpage
\begin{figure}[t!]
    \centering
    \includegraphics[width=\textwidth]{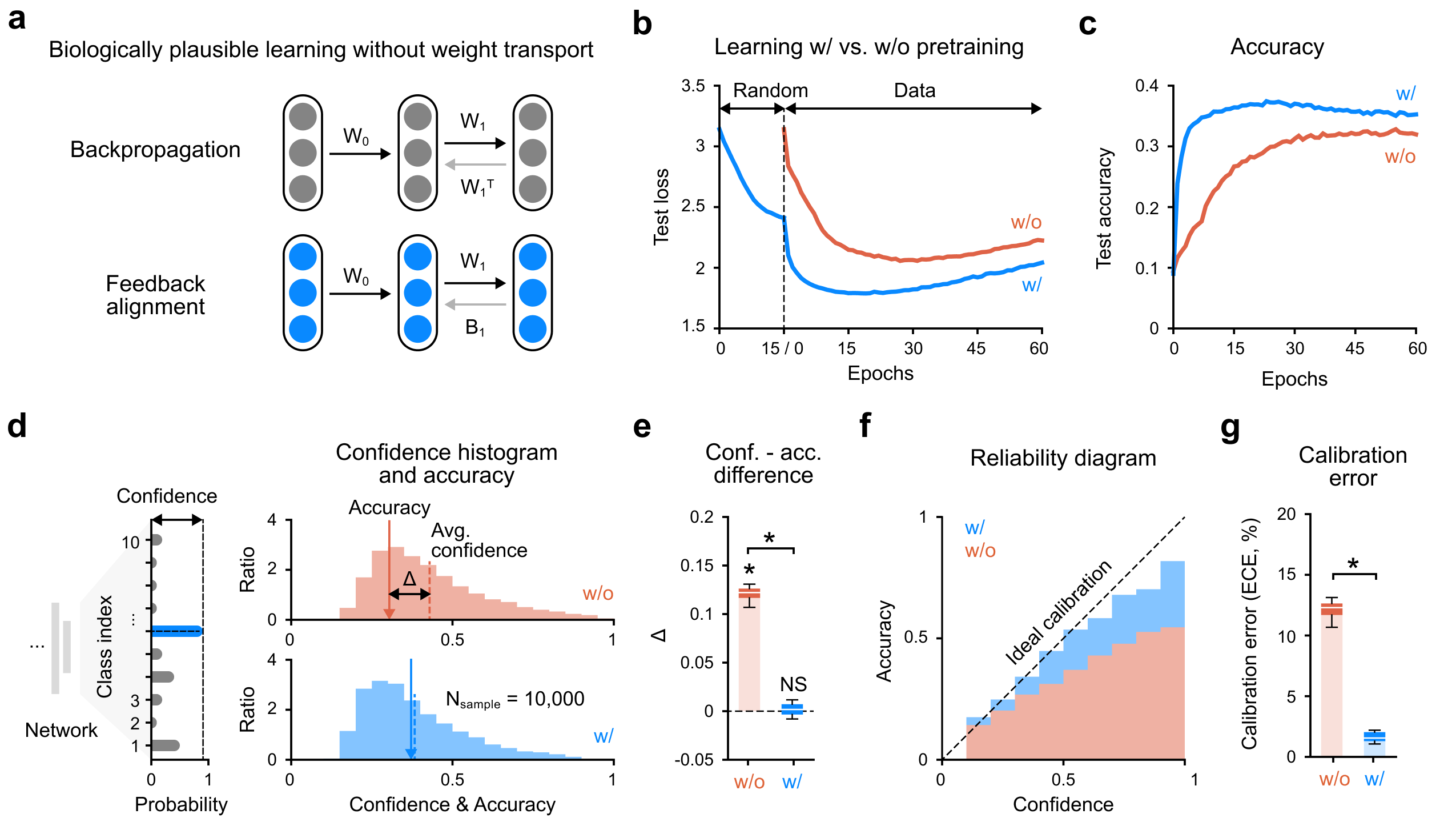}
    \caption{
        \textbf{Uncertainty calibration via random noise pretraining with biologically plausible learning.} \\
        (a) Comparison of backpropagation and feedback alignment. Backpropagation is the standard training method for artificial neural networks but is not considered biologically plausible due to the weight transport problem, which requires a symmetric backward pathway for transmitting error signals. Feedback alignment, on the other hand, is a biologically plausible alternative that does not rely on weight transport, instead using a random matrix B to replace the backward synaptic pathway.
        (b) Test loss during random noise pretraining and subsequent data training. Blue indicates networks pretrained with random noise, and orange indicates networks trained solely with data (without random noise pretraining).
        (c) Test accuracy during subsequent data training.
        (d) Histogram of predictive confidence, with vertical lines indicating the averaged accuracy and confidence.
        (e) The difference between the averaged confidence and accuracy of predictions (w/o vs. w/ random pretraining, $n_{\text{net}} = 10$, Wilcoxon rank-sum test, $P < 10^{-3}$; w/o vs. zero, Wilcoxon signed-rank test, NS, $P < 10^{-3}$; w/ vs. zero, Wilcoxon signed-rank test, NS, $P = 0.752$).
        (f) Reliability diagram showing expected sample accuracy as a function of binned confidence.
        (g) Expected Calibration Error (w/o vs. w/ random pretraining, $n_{\text{net}} = 10$, Wilcoxon rank-sum test, $P < 10^{-3}$).
    }
    \label{figs3}
\end{figure}

\clearpage
\begin{figure}[t!]
    \centering
    \includegraphics[width=\textwidth]{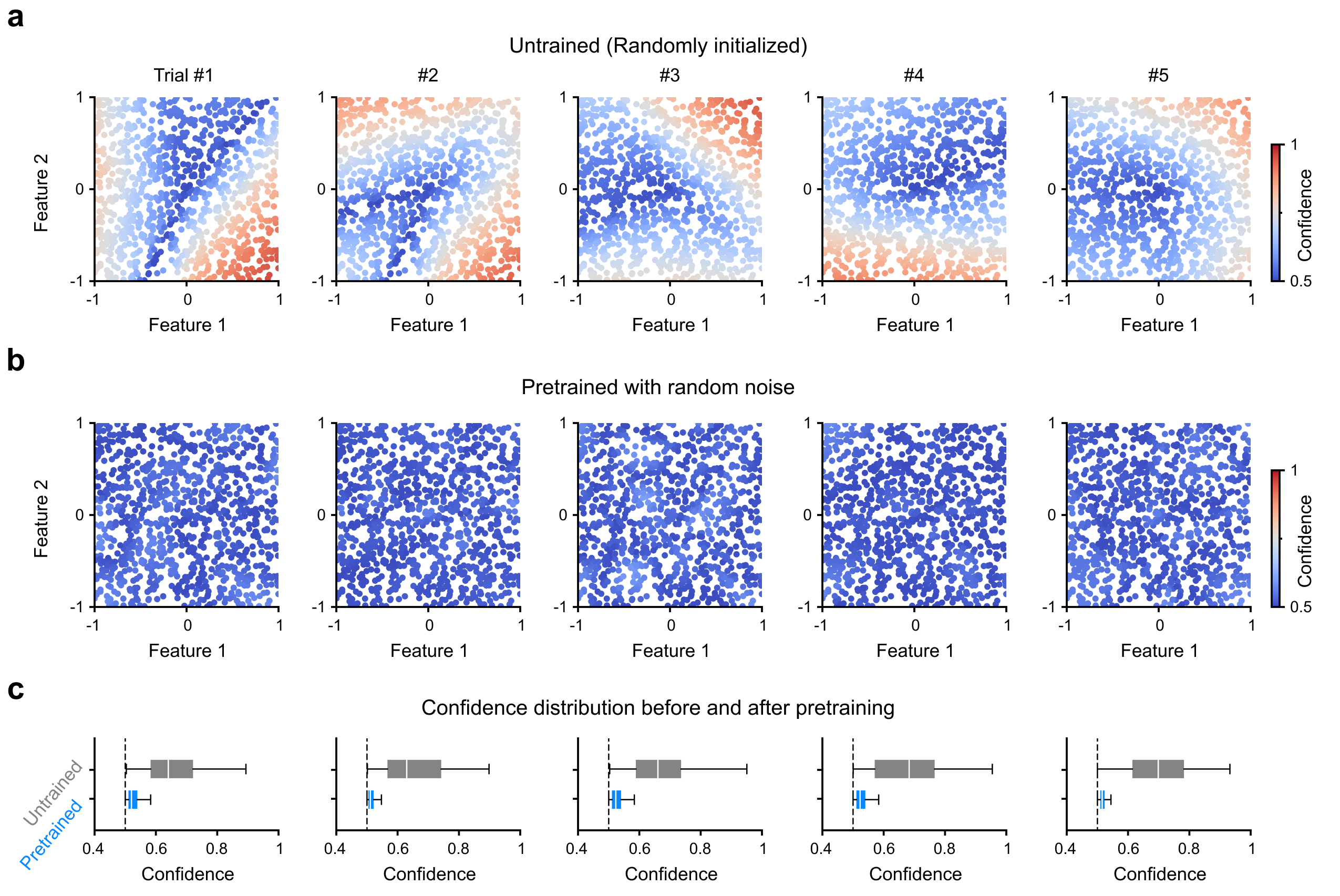}
    \caption{
        \textbf{Pre-calibration of uncertainty in toy-model network.} \\
        (a-c) To visualize the confidence over input space, a toy-model network with a two-dimensional input and output space was employed. Five trials with different random seeds were visualized.
        (a) Visualization of confidence over input space in the untrained network.
        (b) Visualization of confidence over input space in the random noise pretrained network.
        (c) Confidence distribution before and after random noise pretraining.
    }
    \label{figs4}
\end{figure}

\clearpage
\begin{figure}[t!]
    \centering
    \includegraphics[width=\textwidth]{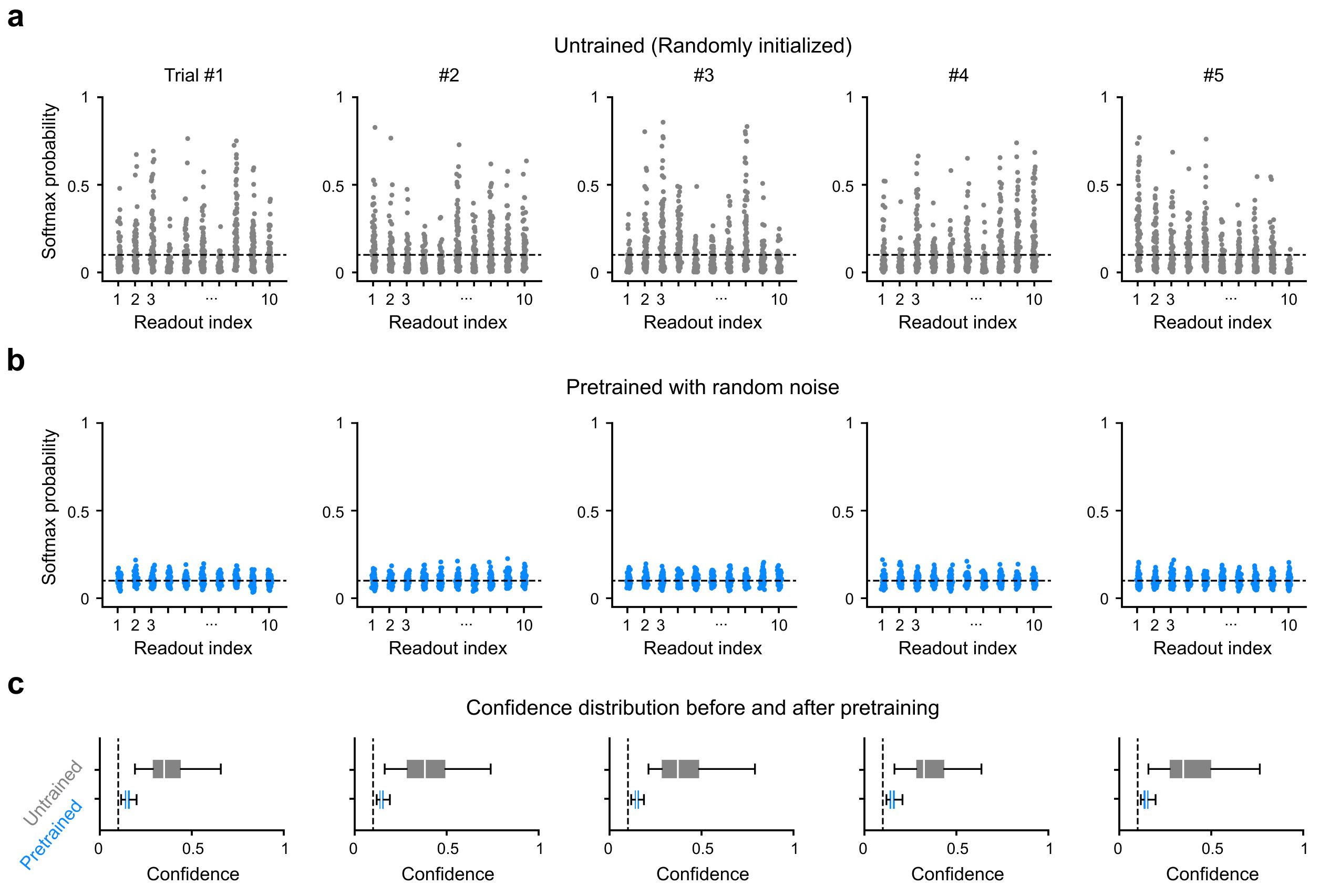}
    \caption{
        \textbf{Pre-calibration of uncertainty in high-dimensional network.} \\
        (a-c) A multi-layer feedforward neural network with a high-dimensional input space was employed to confirm uncertainty calibration via random noise pretraining. Five trials with different random seeds were visualized.
        (a) Visualization of SoftMax output probabilities in the untrained network.
        (b) Visualization of SoftMax output probabilities in the random noise pretrained network.
        (c) Confidence distribution before and after random noise pretraining.
        }
    \label{figs5}
\end{figure}

\clearpage
\begin{figure}[t!]
    \centering
    \includegraphics[width=\textwidth]{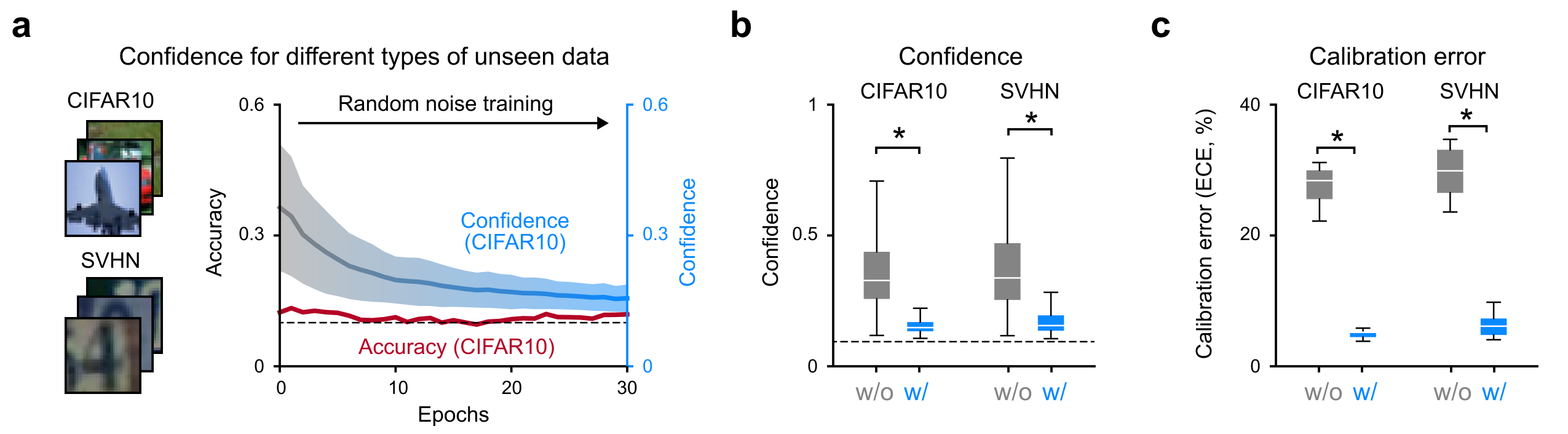}
    \caption{
        \textbf{Calibration of unseen real-world data through random noise pretraining.} \\
        (a-b) Confidence reduction for various unseen real data during random noise pretraining.
        (a) CIFAR-10 and SVHN datasets presented to the network during random noise training, with measured accuracy and confidence (Left). Confidence and accuracy for the CIFAR-10 dataset (Right).
        (b) Confidence distribution for CIFAR-10 and SVHN datasets before and after random noise pretraining.
        (c) Expected Calibration Error (ECE) for CIFAR-10 and SVHN datasets before and after random noise pretraining.
    }
    \label{figs6}
\end{figure}

\end{document}